\documentclass{article}

\usepackage[preprint]{neurips_2023}

\usepackage{times}
\usepackage{soul}
\usepackage{url}
\usepackage[utf8]{inputenc} %
\usepackage[T1]{fontenc}    %
\usepackage{url}            %
\usepackage{nicefrac}       %
\usepackage[hidelinks]{hyperref}
\usepackage[small]{caption}
\usepackage{svg}
\usepackage{graphicx}
\usepackage{amsmath}
\usepackage{amsthm}
\usepackage{array}
\usepackage{multirow}
\usepackage{supertabular,booktabs}
\usepackage{algorithm}
\usepackage{algorithmic}
\usepackage[flushleft]{threeparttable}
\usepackage{nicefrac}       %
\usepackage{amsfonts}       %
\usepackage{microtype}      %
\usepackage{xcolor}         %
\usepackage{listings}
\usepackage{todonotes}
\usepackage{tabularx}
\usepackage{wrapfig}
\usepackage{longtable}
\usepackage{tabularx}
\usepackage{booktabs}

\usepackage[switch]{lineno}

\newcommand\blfootnote[1]{%
  \begingroup
  \renewcommand\thefootnote{}\footnote{#1}%
  \addtocounter{footnote}{-1}%
  \endgroup
}

\usepackage{caption} 
\newcommand{\megawika}{MegaWika}

\title{MegaWika: Millions of reports and their sources across 50 diverse languages}

\author{
    Samuel Barham$^{\heartsuit}$
    \And
    Orion Weller$^{\heartsuit}$
    \And
    Michelle Yuan$^{\spadesuit}$
     \And
    Kenton Murray$^{\heartsuit}$
    \And
    Mahsa Yarmohammadi$^{\heartsuit}$
    \And
    Zhengping Jiang$^{\heartsuit}$
    \And
    Siddharth Vashishtha$^{\diamondsuit}$
    \And
    Alexander Martin$^{\diamondsuit}$
    \And
    Anqi Liu$^{\heartsuit}$
    \And
    Aaron Steven White$^{\diamondsuit}$
    \And
    Jordan Boyd-Graber$^{\clubsuit}$ 
    \And
    Benjamin Van Durme$^{\heartsuit}$\\
    \and
    Human Language Technology Center of Excellence \\
   Johns Hopkins University\\
   \texttt{samuel.barham@jhuapl.edu oweller2@jhu.edu vandurme@jhu.edu}
}

\newif\ifcomment\commenttrue

\usepackage[a-1b]{pdfx}

\usepackage{framed}
\usepackage{mdwlist}
\usepackage{siunitx}
\usepackage{latexsym}
\usepackage{colortbl}
\usepackage{xcolor}
\usepackage{nicefrac}
\usepackage{booktabs}
\usepackage{fnpct}
\usepackage{amsfonts}
\usepackage[T1]{fontenc}
\usepackage{bold-extra}
\usepackage{amsmath}
\usepackage{amssymb}
\usepackage{bm}
\usepackage{graphicx}
\usepackage{mathtools}
\usepackage{microtype}
\usepackage{multirow}
\usepackage{multicol}
\usepackage{xpatch}
\usepackage{latexsym,comment}
\usepackage[normalem]{ulem}

\newcommand*{\missingreference}{{\Huge \colorbox{red}{?reference?}}}
\newcommand*{\missingcitation}{{\Huge \colorbox{red}{?citation?}}}

\makeatletter
\xpatchcmd{\@setref}{\bfseries}{\missingreference}{}{}
\def\@citex[#1]#2{\leavevmode
    \let\@citea\@empty
    \@cite{\@for\@citeb:=#2\do
        {\@citea\def\@citea{,\penalty\@m\ }%
            \edef\@citeb{\expandafter\@firstofone\@citeb\@empty}%
            \if@filesw\immediate\write\@auxout{\string\citation{\@citeb}}\fi
            \@ifundefined{b@\@citeb}{\hbox{\reset@font\missingcitation}%
                \G@refundefinedtrue
                \@latex@warning
                {Citation `\@citeb' on page \thepage \space undefined}}%
            {\@cite@ofmt{\csname b@\@citeb\endcsname}}}}{#1}}
\makeatother

\newcommand{\gem}[1]{\mbox{\textsc{gem}}}
\newcommand{\abr}[1]{\textsc{#1}}
\newcommand{\camelabr}[2]{{\small #1}{\textsc{#2}}}

\newcommand{\hidetext}[1]{}
\newcommand{\ignore}[1]{}

\ifcomment
    \newcommand{\pinaforecomment}[3]{\colorbox{#1}{\parbox{.8\linewidth}{#2: #3}}}

    \newcommand{\prtodo}[1]{\pinaforecomment{lightblue}{pr}{#1}}
    \newcommand{\prtodoi}[1]{\pinaforecomment{lightblue}{pr}{#1}}
\else
    \newcommand{\pinaforecomment}[3]{}
    \newcommand{\prtodo}[1]{}
    \newcommand{\prtodoi}[1]{}
\fi

\newcommand{\smallurl}[1]{ \begin{tiny}\url{#1}\end{tiny}}

\definecolor{lightblue}{HTML}{3cc7ea}
\definecolor{CUgold}{HTML}{CFB87C}
\definecolor{grey}{rgb}{0.95,0.95,0.95}
\definecolor{ceil}{rgb}{0.57, 0.63, 0.81}
\definecolor{UMDred}{HTML}{ed1c24}
\definecolor{UMDyellow}{HTML}{ffc20e}

\newcommand{\qa}[0]{\abr{qa}}

\newcommand{\triviaqa}{\camelabr{Trivia}{qa}}

\newcommand{\squad}{\textsc{sq}{\small u}\textsc{ad}}
\newcommand{\nq}[0]{\abr{nq}}

\newcommand{\bert}{\abr{bert}}

\newcommand{\tydi}[0]{\abr{t}{\small y}\abr{d}{\small i}\abr{qa}}

\graphicspath{ {./}{auto_fig/} }

\begin{document}
\maketitle
\blfootnote{$^{\heartsuit}${\text {Johns Hopkins University}}\ \ $^{\clubsuit}${\text{University of Maryland, College Park}}  \ \ $^{\diamondsuit}${\text{University of Rochester}}}

 \blfootnote{$^{\spadesuit}${\text{Amazon, work completed while at UMD}}}

\begin{abstract}

 To foster the development of new models for collaborative AI-assisted report generation, we introduce \megawika{}, consisting of 13 million Wikipedia articles in 50 diverse languages, along with their 71 million referenced source materials. We process this dataset for a myriad of applications, going beyond the initial Wikipedia citation extraction and web scraping of content, including translating non-English articles for cross-lingual applications and providing FrameNet parses for automated semantic analysis. \megawika{} is the largest resource for sentence-level report generation and the only report generation dataset that is multilingual. We manually analyze the quality of this resource through a semantically stratified sample. Finally, we provide baseline results and trained models for crucial steps in automated report generation: cross-lingual question answering and citation retrieval.
 
\end{abstract}

\section{Introduction}
\label{sec:introduction}

There is a surge in popular demand for collaborative AI based on large language models, such as in the authoring of new documents. In this work we introduce a resource meant to foster the development of collaborative authoring of \emph{reports} based on \emph{multilingual sources} of information. This resource, \megawika{}, is constructed from the largest open, collaborative report authoring dataset in the world, Wikipedia.
\megawika{} comprises more than 13 million Wikipedia articles across 50 languages. To accomplish this, Wikipedia passages and referenced web source materials are extracted, automatically translated into English, semantically analyzed, and their source materials are scraped and automatically cleaned. Finally, for each of the 71 million passage/source pairs, questions are extracted, yielding more than 120 million automatically-generated question/answer (\qa{}) pairs. %

Unlike most other similarly structured datasets, where data typically comes only from some homogeneous, well-behaved corpus (such as Wikipedia exclusively, or collections of news text), our Wikipedia context passages are tied each to a related document taken from the Internet as a whole, leading to a collection that is stylistically and structurally diverse. Moreover, as the non-English documents were not generated automatically through machine translation, they may be expected to better resemble corpora targets by real world cross-lingual question answering (XLQA) systems as well as information retrieval (CLIR) applications. \megawika{} also enables high quality model pretraining for many tasks, which we exemplify  in Section~\ref{sec:applications}.

Automatic processes lead to errors, and in addition, it is not guaranteed that sources cited by the author of a Wikipedia article do in fact represent high quality citations. We perform a number of investigations on the quality of the current resource, and describe steps taken to improve on initial Wikipedia extractions.  The full collection is 1.1TB in size, hosted on HuggingFace's dataset repository to allow for easy usage through dataset streaming\footnote{https://huggingface.co/datasets/hltcoe/megawika}.

In summary: 
\begin{enumerate}
    \item We introduce \megawika{}, a naturally cross-lingual dataset consisting of over 120 million English question/answer pairs spread over more than 50 languages, the largest and most diverse resource of its kind. The 50 languages selected for MegaWika were chosen thoughtfully to include examples from a wide range of language families.  %
    \item We provide a novel quantitative analysis of Wikipedia citations' crosslingualism in Section~\ref{sec:analysis}. Previous quantitative analyses of Wikipedia's citation behavior across languages has relied mostly on information that can be deduced from the URLs of the citations themselves (e.g., \cite{analysis-of-references-2017-lewoniewski}); ours relies on scraping and processing a substantial subset of the web citations across the 50 chosen Wikipedias.
    \item We provide a semantic analysis for all Wikipedia content, allowing for structured exploration and semantic filtering.
    \item We release manual annotations reflecting the level of evidential support between a source and Wikipedia-based questions, to enable building models for automatic assessment.
    \item We illustrate the use of \megawika{} in information seeking tasks including cross-lingual QA and citation retrieval, with associated model artifacts released.
\end{enumerate}

\section{Dataset Collection}
\label{sec:dataset-collection}

The collection approach is as follows:
\begin{enumerate}
    \item \textbf{Identify passages} in Wikipedia,  i.e., 1-3 sentences with a trailing external web citation.\footnote{Wiki dumps for each language were downloaded between Mar. 25 of 2022 (English) and Oct. 20, 2022 (Irish); most Wiki dumps were downloaded in April of 2022, which should be considered \megawika{}'s effective knowledge cutoff.} %
    \item \textbf{Scrape} raw HTML from the cited web page, and \textbf{Extract} the human-readable \textbf{content} from the scraped page, discarding elements extraneous to the underlying source document (navigation bars, menus, advertisements, etc.) -- the result we call a \textit{source document}. Here we leverage \href{https://trafilatura.readthedocs.io/en/latest/}{Trafilatura}, an open-source library for web content extraction \citep{barbaresi2021trafilatura}. The process took approximately two months on 11 high-memory bandwidth AWS instances.
    \item In the case of non-English Wikipedia, \textbf{translate} the Wikipedia passage into English: this allows MegaWika to serve both as a monolingual resource in each of 50 languages, as well as a cross-lingual resource centered on English.\footnote{Future work can consider replicating just the translation component atop our effort, in order to build a cross-lingual artifact centered on a non-English language.} 
    \item \textbf{Extract events} using the LOME FrameNet parser \citep{xia-2021-lome}; these events correspond with high likelihood to semantically salient factive information in the passage, and are thus relevant answers to natural questions.
    \item \textbf{Automatically generate questions} based on the English Wikipedia passage (or its  translation). In the case of question/answer pairs generated from a translated passage, following the data projection methods in \cite{yarmohammadi-etal-2021-everything}, we \textbf{align the translated passage} with its non-English original and determine the most likely non-English span corresponding to the English answer.
    \item \textbf{Locate the answers} to those questions as spans in the cited source document; if they are present, we have an \textit{exact answer match} in the source.
\end{enumerate}

\begin{figure*}
   \centering
    \includegraphics[width=1.0\textwidth]{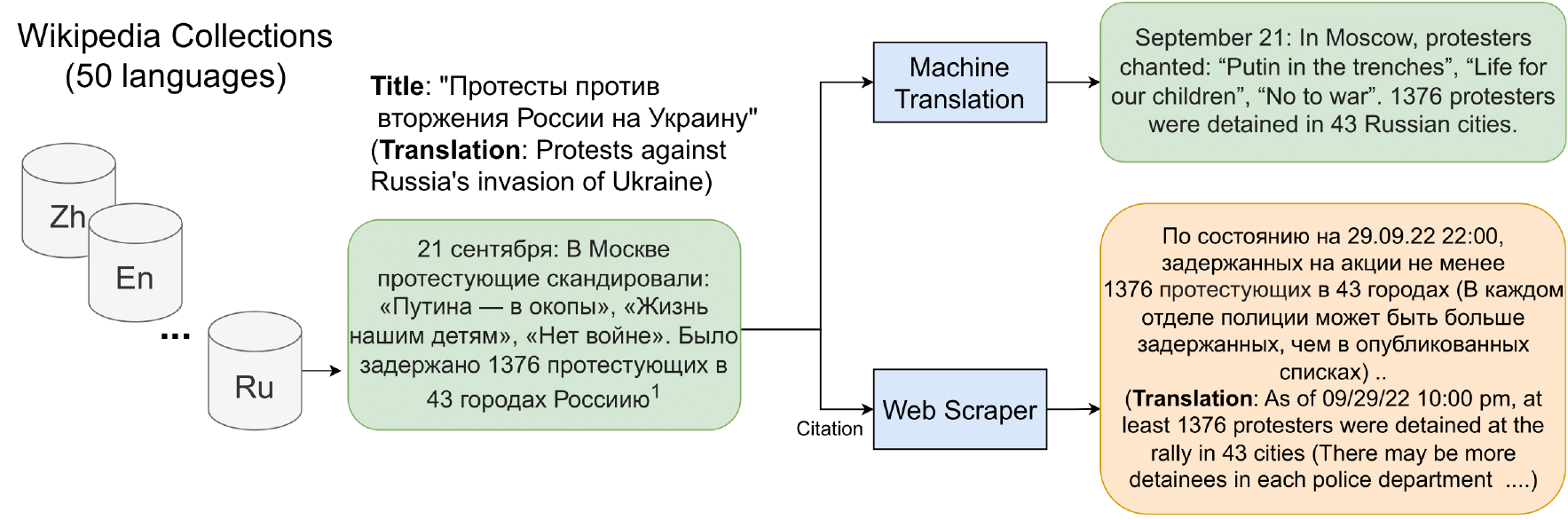}
    \caption{The \megawika{} collection process, illustrated over a Russian Wikipedia example. Wikipedia articles are split into passages with citations, and the original Wikipedia article is translated and the source link is scraped and kept in the original language. Note that there are two additional steps not shown for clarity: question generation from the machine translated Wikipedia article and question-answer span alignment.}
    \label{fig:process_english}
\end{figure*}

In the following we focus on three key aspects of the pipeline\,---\,\textbf{machine translation}, \textbf{question/answer generation}, and \textbf{semantic analysis}.

\paragraph{Translate Passages into English}
For each of the 49 non-English languages, we translate their collected Wikipedia passages into English, storing the results alongside the original language version of the passages. %
Each passage is split into sentences using spaCy, relying on language-specific models where possible \citep{Honnibal_spaCy_Industrial-strength_Natural_2020}. Then, each sentence is translated using M2M-100, a powerful, open-source Machine Translation system that focuses on balancing data for language pairs beyond English, scaling up to 100 languages \citep{fan2021beyond}.\footnote{We use the 418 million-parameter model, with the standard 128k sentence piece tokens, a beam size of 5, and we allow the max number of tokens in a sentence to be clipped to 1,000.} Throughout the dataset collection cycle, we observed that Google Translate often produces higher quality translations, particularly in low-resource languages. As a result, we provide Google translations (not M2M-100 translations) for the lowest frequency 10 languages, and an updated release of \megawika{} will contain dual translation for all languages. Details and statistics of the Google translation data are in the supplementary materials. %

\paragraph{Question-Answer Pair Generation}
We generate questions based on the English versions of Wikipedia passages using \abr{paq}~\citep{lewis-2021}, a system that outputs factoid questions given a document.\footnote{In early development we developed a similar generation model independently, but then embraced the \abr{paq} model to have alignment across these related artifacts based on Wikipedia.} \abr{paq} involves four steps: 1) passage selection, 2) answer extraction, 3) question generator, 4) filtering. The passage selector is a RoBERTa~\citep{liu-2019} model that is trained to select passages with information for factoid questions. The answer extractor is a \bert{}~\citep{devlin2018bert} model that is trained to detect spans of texts that are likely answers to questions.  The question generator is a \abr{bart}~\citep{lewis-2020} model fine-tuned on \nq{}~\citep{kwiatkowski-2019}, \triviaqa{}~\citep{joshi-2017}, and \squad{}~\citep{rajpurkar-16}. This model generates a question conditioned on the selected passage and an extracted answer. The filtering step only keeps questions that are unambiguous. We omit this step here, as we want to include questions that are not directly answered by the Wikipedia article\,---\,but could be answered by the cited source web page. Finally, we record all spans in the source document that exactly match the generated question's answer span.

\paragraph{Semantic Analysis}
To enable semantic-based corpus analysis, we turn to FrameNet~\citep{baker-1998}. A frame is defined as a concept that describes some event, relation, or entity. Each frame is associated with a set of roles, which are triggered by certain spans in the sentence. Each passage is parsed using the \abr{lome} FrameNet parser~\citep{xia-2021-lome}, which predicts which spans of text evokes frames and their associated roles.  %
These annotations enable structured exploration and semantics based sampling of \megawika{} (c.f. the supplemental materials for analysis and statistics).

\section{Dataset Description}
\label{sec:dataset-description}

\paragraph{Structure}
Each entry in \megawika{} comprises a single Wikipedia article, along with a list of all its extracted passages. Each entry in this list is a dictionary containing: the passage text; its machine translation into English (where appropriate); content extracted from the web source its cites; generated question/answer pairs; and the passage's FrameNet analysis. The full hierarchical structure of an individual entry is specified in detail on \megawika{}'s dataset card, which is hosted at \megawika{}'s \href{https://huggingface.co/datasets/hltcoe/\megawika{}}{homepage} on \href{https://huggingface.co/}{HuggingFace}. It is also described in Appendix A of the supplementary materials.

\begin{figure*}[h!]
   \centering
    \includegraphics[trim={3cm 0 0 0},clip,width=\textwidth]{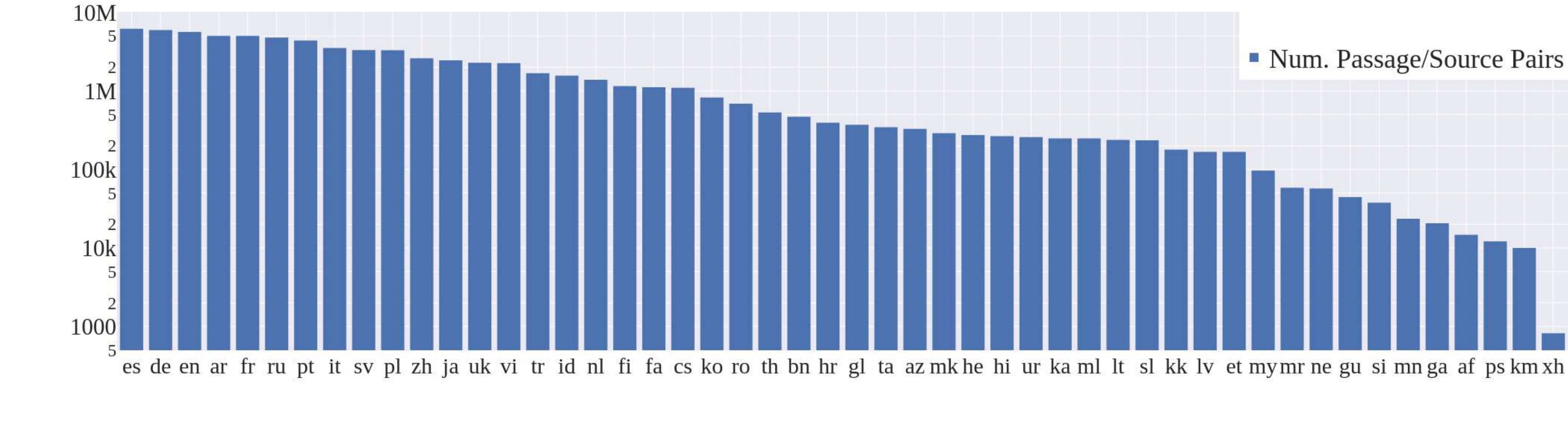}
    \caption{\megawika{} passage/source counts across languages, labeled by ISO 639-1 language codes. The Y-axis is on a log scale. On average, each passage/source pair yields 1.8 question/answer pairs.}
    \label{tab:dataset-size}
\end{figure*}

\paragraph{Statistics}
The current version of \megawika{} spans some 128 million question answer pairs, spread across approximately 71 million context-passage pairs (each consisting of (1) a Wikipedia passage, and (2) the core textual content of the linked web page, cited in the Wikipedia passage, which we call a source document). The Wikipedia passages are drawn from English Wikipedia and 49 of the largest non-English Wikipedias (Figure~\ref{tab:dataset-size}). These 49 languages were selected both for the scale of their Wikipedias as well as to ensure coverage of a diverse set of language families. %

\section{Analysis and Evaluation}
\label{sec:analysis}

\begin{figure*}[h]
   \centering
    \includegraphics[trim={0.75cm 0 0 0},clip,width=\textwidth]{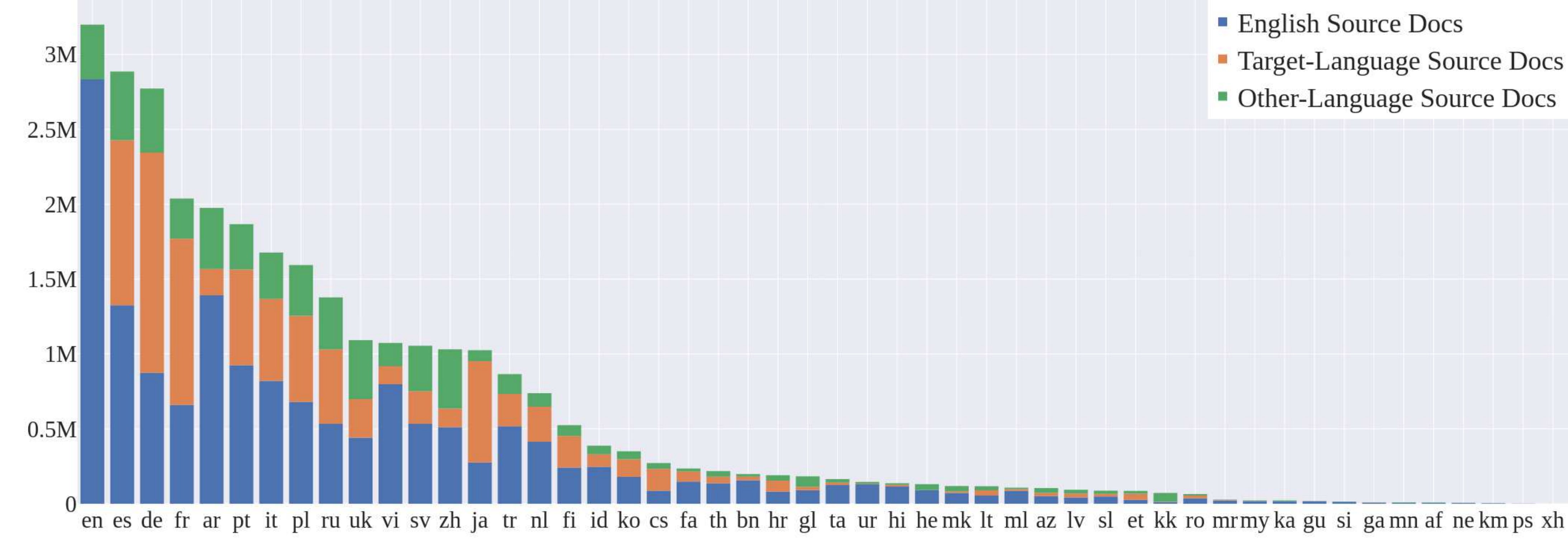}
    \caption{Distribution of the source documents by Wikipedia language, labeled by ISO 639-1 language codes.}
    \label{tab:high-resource-citations}
\end{figure*}

\paragraph{Crosslinguality}
Automatic language identification (LID) using PyCLD2\footnote{Found here: \url{https://pypi.org/project/pycld2/}} revealed two interesting phenomena, both of which illuminate the inherent cross-lingual nature of Wikipedia source citations. First, while English Wikipedia cites non-English sources quite frequently, a full 11\% of online source citations -- more than 2 million citations -- were to non-English web documents; these web documents span a great diversity of languages, running the gamut from high resource languages to very low resources languages. Second, across the 49 non-English Wikipedias, the \emph{majority} of online source citations were to languages \emph{other} than the Wikipedia’s native language. In fact, 48\% were to English web sources, and 19\% were to other languages (i.e., other than the language of that Wikipedia version). Only the remaining 33\% of citations were to documents in the same language as the Wikipedia passage itself. This phenomenon was found to be most concentrated in low-resource languages: for example, Xhosa, Pashto, and Khmer citations are nearly 90\% English; in fact, Xhosa cites no Xhosa sources at all. This is seen even in some high-resource languages. For example, Arabic Wikipedia only cites Arabic websites 9\% of the time, Chinese Wikipedia only cites Chinese websites 12\% of the time, and Farsi Wikipedia cites Farsi web documents 28.5\% of the time (Figure~\ref{tab:high-resource-citations}).

\begin{figure*}[h!]
   \centering
    \includegraphics[width=1\textwidth]{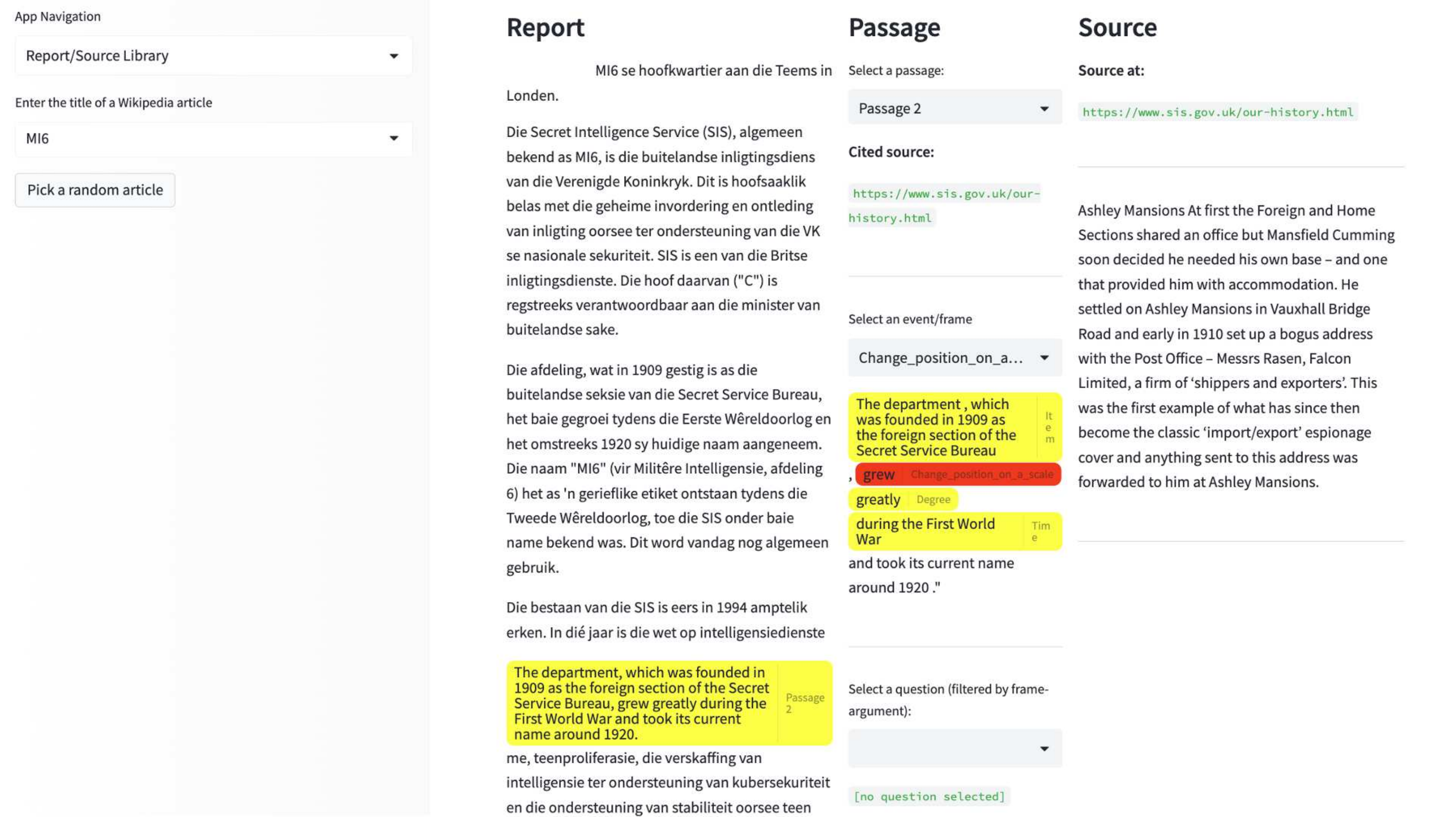}
    \caption{Example of the \megawika{} data viewer; Afrikaans article on MI6 in focus.}
    \label{fig:viewer}
\end{figure*}

\paragraph{Viewer} We provide a viewer for manual exploration, hosted in a HuggingFace space. It allows filtering Q/A pairs by triggered FrameNet frames, and inspection of cited source documents (Figure~\ref{fig:viewer}).

\subsection{Manual Evaluation}
\label{sec:manual-evaluation}

We perform multiple rounds of manual analysis on \megawika{}: (1) we first stratify sample passages according to semantic events, asking crowdworkers to verify when highlighted events are supported by the source; then (2) we provide an author-based analysis of a subset of these examples to determine quality of each step of the pipeline; then (3) we devise a second protocol  for crowdsource annotation that is based on question answering, which we prove out on the subset examined by the authors; finally (4) we collect a development and test set under this protocol, in order to enable future model-based data filtering and automatic scoring of how well content supports a passage.

\paragraph{Semantic Stratified Sampling}

To judge the quality of the source extraction with respect to the passage text, we take a sample of the native English portion of \megawika{}. 
To ensure a diverse mix of types of described situations from the passage-source pairs, we sample passages based on the FrameNet frames predicted within those passages.\footnote{We restrict to \emph{situations} (events, processes, or states), leading to a list of roughly 590 such frames.}
We target 5 sampled passages that evoke each frame type: to achieve this we sample a number of passages for each frame based on the evaluated precision of the parser on that type. We set the number of sampled source documents $D_i$ for frame $f_i$ which has Precision $P_i$ on the FrameNet test set as given by a negative binomial distribution:

\begin{align*}
\operatorname{D_i} = 
\begin{cases}
  \lceil\frac{5}{P_i}\rceil, & \text{if }  count(f_i) >= 10 \\
  \lceil\frac{5}{ P_{avg}}\rceil, & \text{otherwise}
\end{cases}
\end{align*}

\noindent where, $P_{avg}$ is the average precision across all frames in the FrameNet corpus test set. This yields a semantically balanced subset of 3,504 documents which we then use for human evaluation.

\paragraph{Leveraging Frames to Judge Source Document Quality}
Is the information specified in each Wikipedia passage actually contained in a cited source document?  
For each of our 3,504 passages we highlight the textual span corresponding to the frame for which the passage was sampled. Given the highlighted span we ask annotators (with 3-way redundancy) the following question: \textit{Does the source contain the exact same event highlighted in the passage?}

We use the majority prediction to determine whether the source contains the situation highlighted in the passage text and find that 48.2\% of the scraped source texts contain the event highlighted in the Wikipedia passage text. This suggests that approximately half of the events mentioned in a passage can be explicitly traced back to the cited source.

\begin{wraptable}{r}{6cm}
\small
    \caption{\label{tab:human-evaluation}\centering{Author evaluation results.}}
    \centering 
    \begin{threeparttable}
        \begin{tabular}{m{0.2\textwidth} m{0.1\textwidth} m{0.03\textwidth}}
     \toprule
        Category  & Avg.  &  $\kappa$ \\ %
        \midrule
        Passage extraction       & 4.47 / 5  & 0.25   \\
        Source scrape            & 3.96 / 5  & 0.38   \\
        Question fluency         & 4.43 / 5  & 0.25   \\
        Quest. reasonableness  & 3.84 / 5  & 0.20   \\
        \midrule
        Answerability/Wiki       & 2.38 / 3  & 0.39   \\
        Answerability/Source     & 1.67 / 3  & 0.43   \\
        Answer correctness       & 2.30 / 3  & 0.37   \\ 
        \bottomrule
        \end{tabular}
    \end{threeparttable}
\end{wraptable}

\paragraph{QA Evaluation}
We then consider the source as a basis for answering questions. We conducted an evaluation on a random subsample of 150 entries from the same document sample, assessed by the authors themselves. We assessed: (1) the quality of the passage extraction, (2) the quality of the source document scrape, (3) the fluency of the generated question, (4) the reasonableness of the question, (5) the answerability of the question given the Wiki passage, (6) the answerability of the question given the source document, and (7) the correctness of the selected answer span. See Table~\ref{tab:human-evaluation}, with inter-annotator agreement according to Fleiss's $\kappa$. We have fair to moderate agreement across all categories of evaluation. We note that roughly half  of the sources in this analysis do not provide clear support to information in the Wikipedia passage (answerability given the source: 1.67/3 = 55.67\%) which resembles our finding of 48.23\% of sources supporting a highlighted event in the previous analysis, based on FrameNet and crowdsourced assessment.

\begin{table*}[h]
\small
    \centering
    \caption{Examples scored at different levels of answerability from the QA Evaluation, with the most relevant supporting evidence from the cited source documents provided here as illustration.}
    \begin{tabular}{p{.2\textwidth}>{\centering\arraybackslash}p{.14\textwidth}>{\centering\arraybackslash}p{.06\textwidth}p{.45\textwidth}}
  \toprule
  \multicolumn{1}{c}{\centering Question} & \multicolumn{1}{c}{\centering Answer} & \multicolumn{1}{c}{\centering Score} & \multicolumn{1}{c}{\centering Most Relevant Source Text} \\
    \midrule
        {\par\ \par What color changes the sticker when it detects ethylene?} & \par\ \par\ \par white to blue & \par\ \par\ \par 3 & A marker on Riley's RediRipe stickers detects a chemical called ethylene gas, which is released by fruit or vegetables as they ripen. As that happens, the sticker turns from white to blue.\\
    \midrule
        What did the international media want from dennis pozniak? & \par\ \par interviews & \par\ \par 2 & In a day fraught with anxiety declining 8 interviews from various Radio, Print, and TV reporters one TV station wouldn't take "NO" for an answer.\\
    \midrule
        {\ \newline What is cramond island made of?} & \par\ \par dolerite & \par\ \par1 & Geologically, Craigleith is a laccolith, a dome-shaped igneous intrusion, composed of essexite, a rock popular for the manufacture of curling stones.\\
\bottomrule
\end{tabular}
\end{table*}

\begin{figure}[h!]
 \centering
 \includegraphics[width=0.9\textwidth]{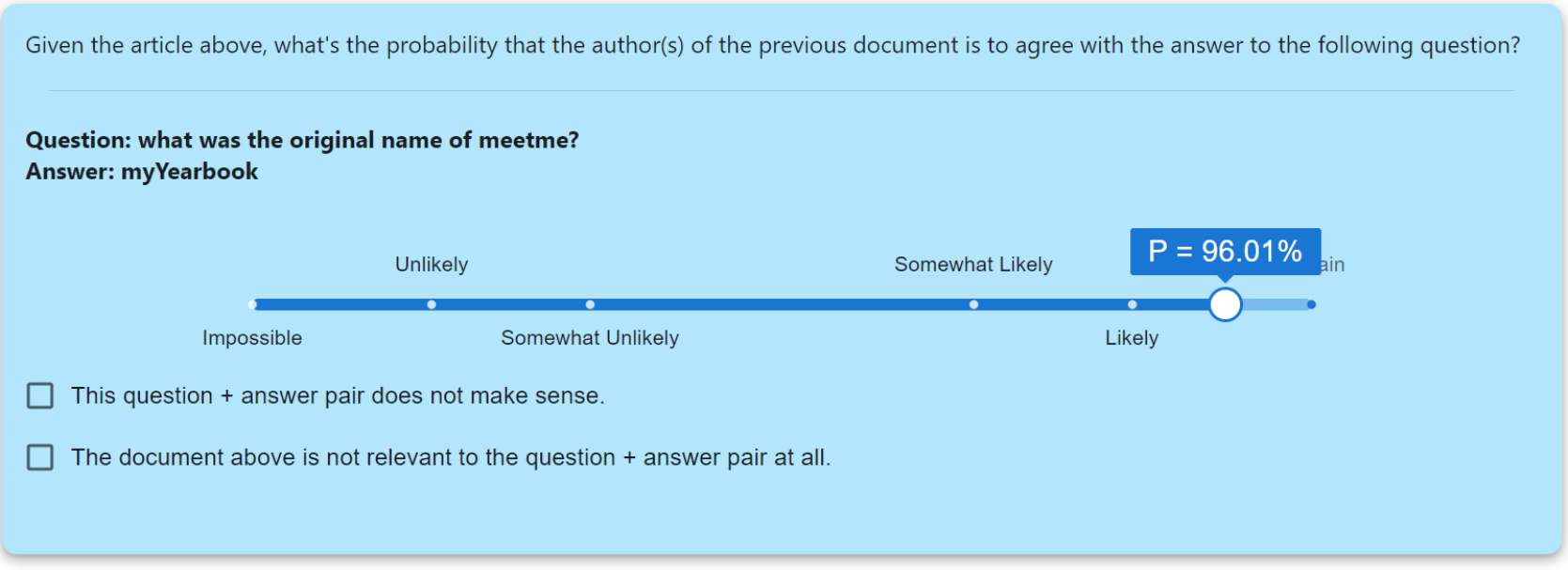}
\caption{Interface for annotating evidential support.}
\label{fig:evidence-ux}
\end{figure}

\paragraph{Strength of Evidential Support Annotation}
Finally, for high-quality question-answer pairs from the previous QA evaluation, we further annotate the scalar \emph{strength} of evidential support from their corresponding source document. %
We extend a previous annotation protocol for Uncertain Natural Language Inference (UNLI) \citep{chen-etal-2020-uncertain} in order to collect fine-grained labels. Crowdsource workers were presented with the interface shown in Figure~\ref{fig:evidence-ux}, paired with a source article.\footnote{We recruit qualified workers with Amazon Mechanical Turk, who achieved exceptional holdout correlation; Turkers were paid \$1 per hit (with 5 instances), which amounts to an hourly wage of roughly \$15.}

\begin{figure*}[h!]
    \begin{minipage}{.6\textwidth}
        \centering
        \includegraphics[width=\linewidth]{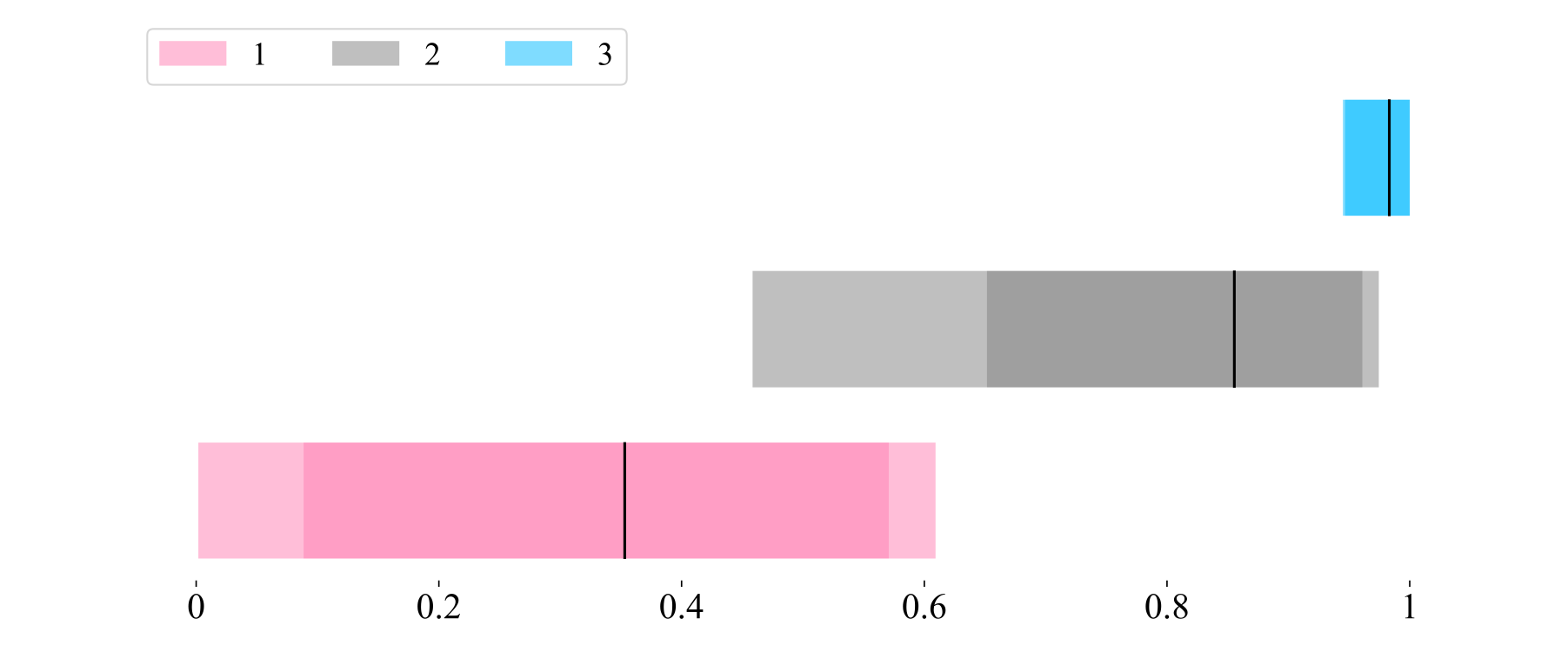}
    \end{minipage}
    \begin{minipage}{.4\textwidth}
        \centering
        \includegraphics[width=1\linewidth]{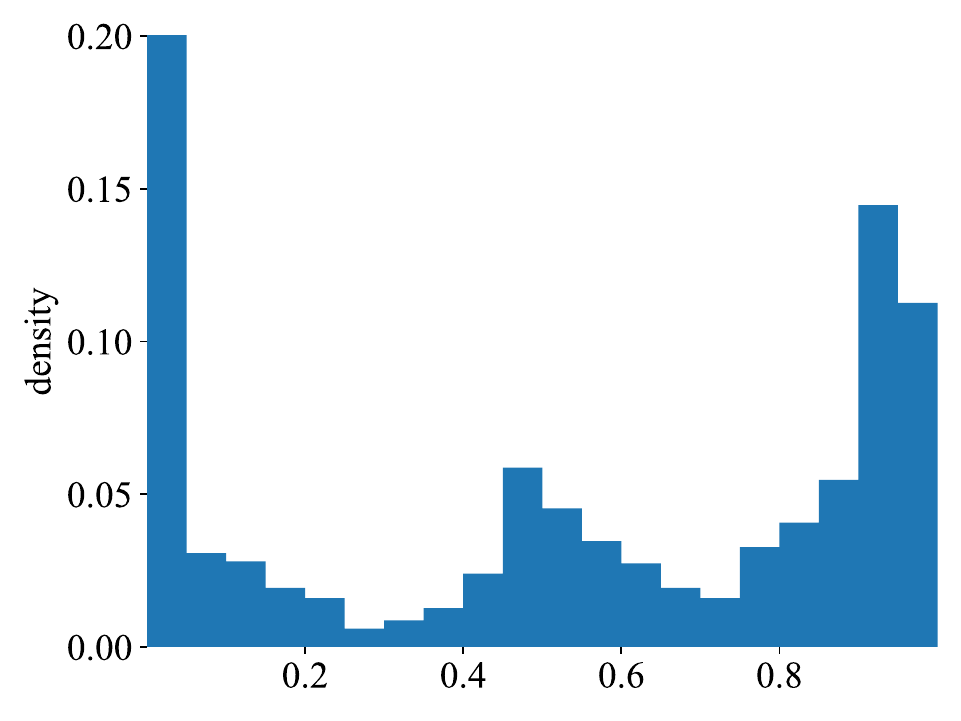}
    \end{minipage}
    \caption{Left: Strength of Evidential Support label distribution for each of the three answerability levels from previous QA evaluation (3 – indisputably answerable and the answer is correct, 1 – completely unanswerable). Light / dark shade covers datapoints within 1 / 1.5 IQR of each category, and the black bar denotes the median. Right: Strength of Evidential Support label distribution on the extended 1.5K English validation set.}
    \label{fig:mega-wika-scalar-label-dist}
\end{figure*}

The left hand side of \autoref{fig:mega-wika-scalar-label-dist} shows that the strength of support from source documents in the \megawika{} dataset has spread across levels of support: sources more or less support statements made in Wikipedia. As this analysis aligns with our in-house answerability annotation, we proceed to create a larger evaluation set that focuses on high quality questions and supporting source documents to support future model based scoring of sources. We filter instances that contain citation-like or table-like text (content with many ``|" and ``-" tokens), sampled 2.5k instances and crowdsource annotated them in a similar way, with 2-way redundancy. These constitute a larger evaluation set with 1.5k validation and 1k test. The distribution of labels from the validation set is on the right hand side of \autoref{fig:mega-wika-scalar-label-dist}. 380 instances were determined by least one annotator as ``unanswerable'' through either or both of the two checkboxes showcased in \autoref{fig:evidence-ux}: these we assigned a supporting strength of 0.\footnote{Of these 380 instances, 207 are marked as ``does not make sense'' by at least one of the two annotators, and 199 are marked as ``irrelevant'' by at least one of the two annotators.}

\section{Example Applications}
\label{sec:applications}
\megawika{} is intended to enable development of models for assisting authors of reports, such as Wikipedia writers needing to locate salient information for their articles. We illustrate two example tasks for finding information: (1) cross-lingual question answering (QA), and (2) citation retrieval.

\subsection{Multilingual Question Answering}
\label{sec:qa}

\megawika{} contains more examples than any previous multilingual QA dataset, including XQuAD  (11 languages, \cite{artetxe2019xquad}), TyDiQA (11 languages, \cite{clark2020tydi}), or MKQA (26 languages,  \cite{longpre2021mkqa}). Furthermore, unlike MKQA and XQuAD, our passages are found naturally on Wikipedia and are not translated from English for research purposes. To demonstrate \megawika{}'s effectiveness for multilingual question answering, we evaluate on XQuAD, subsetting our corpus to only contain the same 11 languages. %

\paragraph{Experiment Settings} In order to gather a high quality subset of the \megawika{} data, we first filter the data to contain only (question, passage, answer) pairs where the answer can be found with exact string match in the passage. Note that this may exclude some aliases, but in this analysis we focus on high precision and leave higher recall for future work.

\begin{table}[t]
\caption{Selecting high-quality question answering data from \megawika{} through self-filtering, scored using exact-match on the XQuAD test set with XLM-R base. We see that one can take \megawika{} and sub-select high-quality examples for cross-lingual question answering.\label{tab:qa}}

\centering
\small
\begin{tabular}{cc|ccc}
\toprule
XQuAD Zero-Shot & XQuAD Translate-Train & Round 1 & Round 2 & Round 3 \\ 
\midrule
65.5 & 77.0 & 73.1 & 76.6 & 78.1 \\
\bottomrule
\end{tabular}
\end{table}
We then conduct several rounds of self-filtering, repeating the following process: (1) we train XLM-R \citep{conneau2019unsupervised} and mBERT \citep{devlin2018bert} on the data until performance plateaus, using 10k sampled instances as a validation set. (2) We use those trained models to filter the data, keeping only instances which are correctly predicted by either of our two trained models or an English DistilBERT trained on English SQuAD \citep{sanh2019distilbert}. (3) We recompile the data into the next dataset and repeat the process. After each iteration we sample 500k instances per language from the dataset to provide the training data for the next iteration of the QA models (except for DistilBERT).

\paragraph{Results} We use the final XLM-R base model from the previous step and evaluate following the translate-train setup in XQuAD \citep{artetxe2019xquad}. %
Our results (Table~\ref{tab:qa}) show that each round of  filtering produces a higher quality model; after several rounds of filtering, it has higher performance to training on XQuAD.  This supports the claim that \megawika{} can be used effectively towards building high-quality multilingual QA systems. Furthermore, in this example we limited our experiments to the 11 languages that are used in XQuAD, but \megawika{} also exists for 39 other languages, including many low-resource languages where this data will be particularly valuable.

\subsection{Multilingual Information Retrieval for Citations}

Another example application of \megawika{} is for large-scale multilingual information retrieval. %

\paragraph{Experiment Settings} We first filter the data to avoid train/test leakage and improve dataset quality (similar to Section~\ref{sec:qa}). Like in our filtering during manual evaluation for evidentiary support, we start by removing all instances that contain citation-like or table-like text, as indicated in Wikipedia by many ``|" and ``-" tokens. As some Wikipedia pages may have similar text (and sources) across languages, we select instances for the evaluation sets, and then remove all other instances from the corpus that are linked through Wikipedia's language links. As the size of the data is extremely large, removing these instances from the evaluation data has only a minor effect on corpus size and allows us to avoid data leakage from similar texts in a different language.

We follow this selection process to gather 1k instances per language, or as much as available, and 10k of English per evaluation set and 50k instances per language for training. We use the set of all source documents as the retrieval collection, consisting of 72M passages.

As the number of languages in \megawika{} far exceeds any previous retrieval models, the closest baseline is the multilingual mDPR model used in \cite{miracl}, although it was used for multilingual retrieval rather than cross-lingual retrieval. Note that other cross-lingual models are at most trained on 5 languages, making them a poor comparison.  We note that mDPR from MiRACL was trained on XOR QA \citep{asai2020xor} (which was adapted from TyDiQA \citep{clark2020tydi}). To allow meaningful comparison, we use the same base model and training process as that in MiRACL, simply changing the training data.

\paragraph{Results} Table~\ref{tab:ir} shows that our citation-finding version of mDPR greatly outperforms the baseline, by around 300\%. We further note that precision at 1000 (the cutoff typically used for re-ranking) is at 48.6. Our citation mDPR also includes a much wider set of languages compared to any previous open-source dense retrieval model, with 50 languages used in training compared to 18 in MiRACL. 

While these experiments are meant as baseline illustrations of what \megawika{} will enable, these models already may provide use to other researchers and will be released along with the data resource.

\begin{table}[t]
\caption{Results for multilingual information retrieval.  Our version of mDPR trained with the same architecture is much more adept at finding sources for citations, as measured by precision (P) and mean reciprocal rank (MRR) at 10. \megawika{} also enables retrieval for 50 languages, which is 2x more than existing IR collections.\label{tab:ir}}

\centering
\small
\begin{tabular}{l|rrrrr}
\toprule
Model & P@5 & P@10 & P@100 & P@1000 & MRR \\ 
\midrule
mDPR (MiRACL) & 5.7 & 6.9 & 11.7 & 18.5 & 4.3 \\
mDPR (Ours) & 18.0 & 21.3 & 33.6 & 48.6 & 14.1 \\
\bottomrule
\end{tabular}
\end{table}

\section{Related Work}
\label{sec:related-work}

\paragraph{Automated Report Generation}
Report generation can have many definitions: \citet{chen2020wikitablet} uses Wikipedia articles and tables for data-to-text report generation, while there also exists a large amount of work on taking medical image data and creating medical reports \citep{li2019knowledge,yang2021writing,liu2021exploring,liu2021auto}. Other work has derived datasets consisting of question-answer pairs for various usages from Wikipedia articles, which we adopted in our question generation approach \citep{lewis-2021}.%

As language models have improved, recent attention has turned towards examining the citations found on Wikipedia as part of report generation. Recent work on this topic has improved the way that these reports cite information through better fact checking \citep{kamoi2023wice,petroni2022improving}.

\cite{qian2023webbrain}, concurrent to this work, is the closest to ours, as they focus on textual report generation in an open-domain setting. They also build a novel dataset, but like \cite{lewis-2021} use English Wikipedia only. There are several other crucial distinctions between our work and theirs: (1) our work scrapes the citations at the sentence level as compared to scraping the citations section and attempting to match them post-hoc, (2) our work aims to enable collaborative report generation, where AI models assist by finding references and suggesting content for a report one portion at a time (while their work provides a Wikipedia article title and asks the model to generate the full article), and (3) although their raw dataset contains more instances, their data available for training is a magnitude smaller than ours (30M English passages in our retrieval collection while they provide 3M).

\paragraph{Using Multilingual Wikipedia}
As Wikipedia is available in many languages, much research has used it for various multilingual applications. Some of these resources (like XQuAD or MKQA \citep{liu-2019,artetxe2019xquad}) use automated or human translations of the English versions, some have used language links \citep{lewis-2020-mlqa} and others have used multilingual  Wikipedia versions with entirely different questions across languages (e.g.  \tydi{} \citet{clark2020tydi}).  

In the information retrieval setting, very few works have used Wikipedia for large-scale multilingual retrieval. \citet{asai2020xor} extended \tydi{} to the cross-lingual open-retrieval question-answering setting (11 languages) while MiRACL extended it further to include 18 languages \citep{miracl}. Our work is different in that we provide information retrieval data for citation-finding, rather than standard web-retrieval. Furthermore, we provide a much larger corpus and include 50 languages.

\section{Conclusions}
\label{sec:conclusion}

We presented MegaWika, a large-scale cross-lingual report generation dataset. 
We described the collection pipeline necessary to construct such a dataset, analyzed the quality of the resource through three distinct human evaluations, and gave examples of ways MegaWika may be used in downstream tasks.  We release this data and associated artifacts through HuggingFace with a custom data browser.

\bibliographystyle{ACM-Reference-Format}
\bibliography{journal-full,neurips23}

 \newpage

 \section*{Appendix A}

Each JSON object comprising the total dataset has the following structure when loaded into a Python dict (for example, using the \texttt{json} module). Python types are specified informally in angle brackets in place of the actual data. Some fields (whose identifiers usually begin with \texttt{lang\_}, referring to non-English material) are irrelevant for the MegaWika split based on English language Wikipedia; in such cases, the fields values are set to the empty string or list, whichever is appropriate.

\colorlet{punct}{red!60!black}
\definecolor{background}{HTML}{EEEEEE}
\definecolor{delim}{RGB}{20,105,176}
\colorlet{numb}{magenta!60!black}

\lstdefinelanguage{json}{
    basicstyle=\normalfont\ttfamily,
    numberstyle=\scriptsize,
    stepnumber=1,
    numbersep=8pt,
    showstringspaces=false,
    breaklines=true,
    frame=lines,
    backgroundcolor=\color{background},
    literate=
     *{0}{{{\color{numb}0}}}{1}
      {1}{{{\color{numb}1}}}{1}
      {2}{{{\color{numb}2}}}{1}
      {3}{{{\color{numb}3}}}{1}
      {4}{{{\color{numb}4}}}{1}
      {5}{{{\color{numb}5}}}{1}
      {6}{{{\color{numb}6}}}{1}
      {7}{{{\color{numb}7}}}{1}
      {8}{{{\color{numb}8}}}{1}
      {9}{{{\color{numb}9}}}{1}
      {:}{{{\color{punct}{:}}}}{1}
      {,}{{{\color{punct}{,}}}}{1}
      {\{}{{{\color{delim}{\{}}}}{1}
      {\}}{{{\color{delim}{\}}}}}{1}
      {[}{{{\color{delim}{[}}}}{1}
      {]}{{{\color{delim}{]}}}}{1},
}

\begin{lstlisting}[language=json, caption="Dataset Object Structure"]
{
    "article_title": <string : title of original Wikipedia article>,
    "article_text": <string : text of Wikipedia article>,
    "entries": [
        ...
        {
        "id": <string : passage ID>
    
        # Wiki passage
        "passage": {
            "text": <list of str : text of passage in English (possibly via MT)>
            "parse": <list of str : per sentence JSON dump of FrameNet parse of English passage text>,
            "en_tokens": <list of str : tokenization of passage in English>,
            "lang_tokens": <list of str : tokenization of original non-English passage>,
            "en_lang_token_map": <list of pairs of str : alignment mapping between English and original language token indices>
        },
    
        # MT
        "mt": {
            "original": <string : original language passage>
            "original_sents": <list of string : sentencized original language passage>,
            "translation": <string : machine translation of passage>,
            "translation_sents": <list of string : sentencized machine translation of passage>,
            "translation_probs": <list of float : log prob of machine translation by sentence, where available>,
            "repetitious_translation": <string \in ("true", "false") : automated judgment on whether machine translation is pathologically repetitious>
        },
    
        # Source document
        "source_lang": <string : language ID of predominant language in source document, 2-character ISO code>
        "source_url": <string : URL of the cited web source>
        "source_text": <string : content extracted from the scrape of the source URL>,
                                
        # Question/answer pairs
        "qa_pairs": [
            ...
            {
            "question": <string : passage ID>
            "en_answer": <string : English answer>,
            "lang_answer": <string : aligned original language answer>,
            "frames": [
                ...
                {
                "frame": <string : frame triggered by the question>,
                "argument": <string : detected frame arguments>
                }
                ...
            ],
            "en_matches_in_source": <list of int : start and end index of the English language-answer token(s) in the source document>,
            "en_match_in_passage": <list of int : start and end index of the English language-answer token(s) in the English language translation of the passage>,
            "lang_matches_in_source": <list of int : start and end index of the original language-answer token(s) in the source document>,
            # "lang_match_in_passage": <list of int : start and end index of the original language-answer token(s) in the original language passage>,
            "passage": <list of string : sentencized view of the passage>,
            "en_answer_tokens": <list of string>,
            "match_disambiguated_question": <string : disambiguated version of question obtained by matching pronouns with article title (noisy but often helpful)>
            }
            ...
        ...
        }
    ]
}
\end{lstlisting}
 \newpage
 \section*{Appendix B}

In Table \ref{tab:human-evaluation}, we already provided examples of question answerability. Here, for completeness, we provide examples of other axes of evaluation from the QA evaluation task described in Section \ref{sec:manual-evaluation}.

\begin{table*}[h]
\small
\centering
\caption{Example passage/source pairs scored at different levels of ``passage extraction'' and ``source scrape'' quality in the QA evaluation.}
\begin{tabularx}{\textwidth}{>{\arraybackslash}Xc>{\arraybackslash}Xc}
\toprule
\multicolumn{1}{c}{\centering Passage} & \multicolumn{1}{c}{\centering Passage Quality} & \multicolumn{1}{c}{\centering Source Doc.} & \multicolumn{1}{c}{\centering Source Quality} \\
\midrule
        {\par\ 2010 58th FAMAS Awards German Moreno Youth Achievement Award 2011 3rd PMPC Star Awards for Music New Male Recording Artist Ngiti Gingtong Kabataan Awards Special Citation 2012 32nd Seal of Excellence, Dangal ng Bayan People's Choice Awards Best Inspiring Male Young Celebrity 43rd Guillermo Mendoza Memorial Scholarship Foundation [...] [...]} & \par\ \par\ 1 & \par\ ``Isang malaking karangalan daw para kay Bea Binene na sila ni Jake Vargas ang nahirang na Most Promising Loveteam For Movies And TV sa katatapos na 42nd Guillermo Mendoza Memorial Scholarship Foundation Awards. [...]
& \par\ \par\
5 \\
    \midrule
        {\par\ Future malls Name Location Land area ($m^2$) Floor area ($m^2$) OpeningKCC Mall of Cotabato Sinsuat Avenue corner Parang Road, Cotabato City 110,000 170,000 2023-2024Veranza Cotabato Beside KCC Cotabato, Quezon Avenue corner Parang Road Cotabato City TBA TBA TBA KCC Mall of Iligan Iligan 70,000 TBA TBA} & \par\ \par\ 2 & \par Inorganic Syntheses, Volume 20. The volumes in this continuing series provide a compilation of current techniques and ideas in inorganic synthetic chemistry. Includes inorganic polymer syntheses and preparation of important inorganic solids, syntheses used in the development of pharmacologically active inorganic compounds, small-molecule coordination complexes, and related compounds. Also contains valuable information on transition organometallic compounds including species with metal-metal cluster molecules. All syntheses presented here have been tested.
What people are saying - Write a review [...] & \par\ \par\
3 \\
\midrule
        {\par\ And as mind-terma are "not physically discovered but are revealed through the mind of the terton," Joseph Smith's revelations of the prophecies of Enoch and the parchment of John did not have any direct physical source but were revealed through Smith's mind. [...]} & \par\ \par\ 5 & \par "Origins. What is religious criticism? ; Enthusiasm, Gnosticism, American optimism ; Cane Ridge through Billy Graham -- American original : the Mormons. A religion becomes a people : the kingdom of God ; The religion-making imagination of Joseph Smith ; Baptism for the dead, spirits for the unborn -- Rival American originals. Christian science : the fortunate fall in Lynn, Massachusetts ; [...] & \par\ \par\ 1 \\
\bottomrule
\end{tabularx}
\end{table*}

\begin{table*}[h]
\small
    \centering
    \caption{Examples question/answer pairs scored at different levels of ``question fluency'' and `question reasonableness'' in the QA evaluation.}
    \begin{tabular}{p{.4\textwidth}>{\centering\arraybackslash}p{.2\textwidth}>{\centering\arraybackslash}p{.1\textwidth}>{\centering\arraybackslash}p{.1\textwidth}}
  \toprule
  \multicolumn{1}{c}{Question} & \multicolumn{1}{c}{Answer} & \multicolumn{1}{c}{\centering Fluency} & \multicolumn{1}{c}{\centering Reasonableness} \\
    \midrule
        what did vladimir uskhopchik command in january 1991? & Vilnius garrison & 5 & 5\\
    \midrule
        what did the supreme court reinstate the two teachers? & imprisonment & 2 & 2 \\
    \midrule
        p.g. wodehouse p.v. 220 what number was not prosecuted & 346 & 1 & 1 \\
\bottomrule
\end{tabular}
\end{table*}

 \newpage
 \newpage
 \section*{Appendix C} \label{sec:appendix-c}

\subsection*{Selection of frames for Human Evaluation}
\label{sec:appendix-c-situation-frames}

We consider 3 categories of frames to denote \textit{situations} - events, processes, or states, as defined by the frame categorization in the FrameNet corpus \footnote{\url{https://framenet.icsi.berkeley.edu/fndrupal/FrameLatticeList}}.
To compile a list of all \textit{situational} frames, we use the following steps:
\begin{itemize}
    \item We create an initial list of frames by tracing the relational trees of 3 frames -- Event, State, and Process. We consider three relations to compile the list, namely Inheritance (child), Subframe (component), and Precedes (earlier/later). This gives us 384 frames.
    \item In addition to categorized frames, FrameNet lists smaller trees that are not classified into any of the categorizations. To collect \textit{situational} frames from this section, two authors of this paper manually select the smaller tree tops that they think denote a situation. For all these selected tree tops, the same three relations described in Step 1 are traced to get a list of 204 additional frames.
\end{itemize}
This gives us a list of 588 frames which are then used to sample passages that contain these frames.

\subsection*{Source Document Quality Evaluation}
\label{sec:appendix-c-source-validation}
\begin{figure}[ht]
    \centering
    \includegraphics[width=\textwidth]{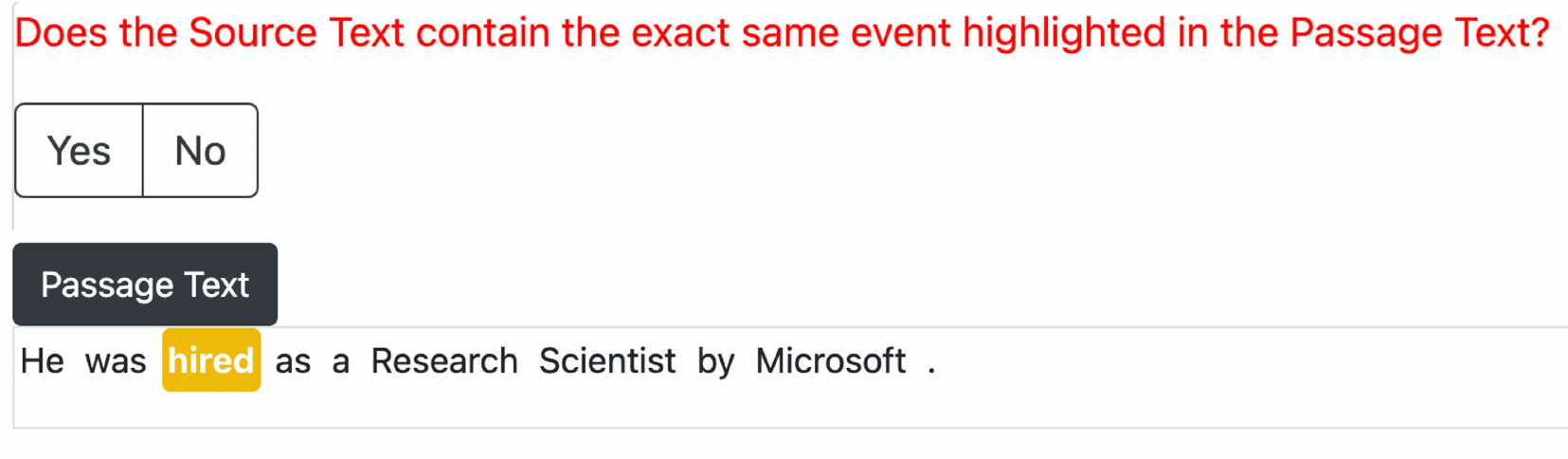}
    \caption{An example of a passage in the annotation protocol used for source validation. The annotators are shown the passage text with a highlighted situation trigger and can then switch to the source text (\autoref{fig:source-val}). This example is a constructed example used in the instructions of the annotation protocol and not from MegaWika.}
    \label{fig:passage-val}
\end{figure}
\begin{figure}[ht]
    \centering
    \includegraphics[width=\textwidth]{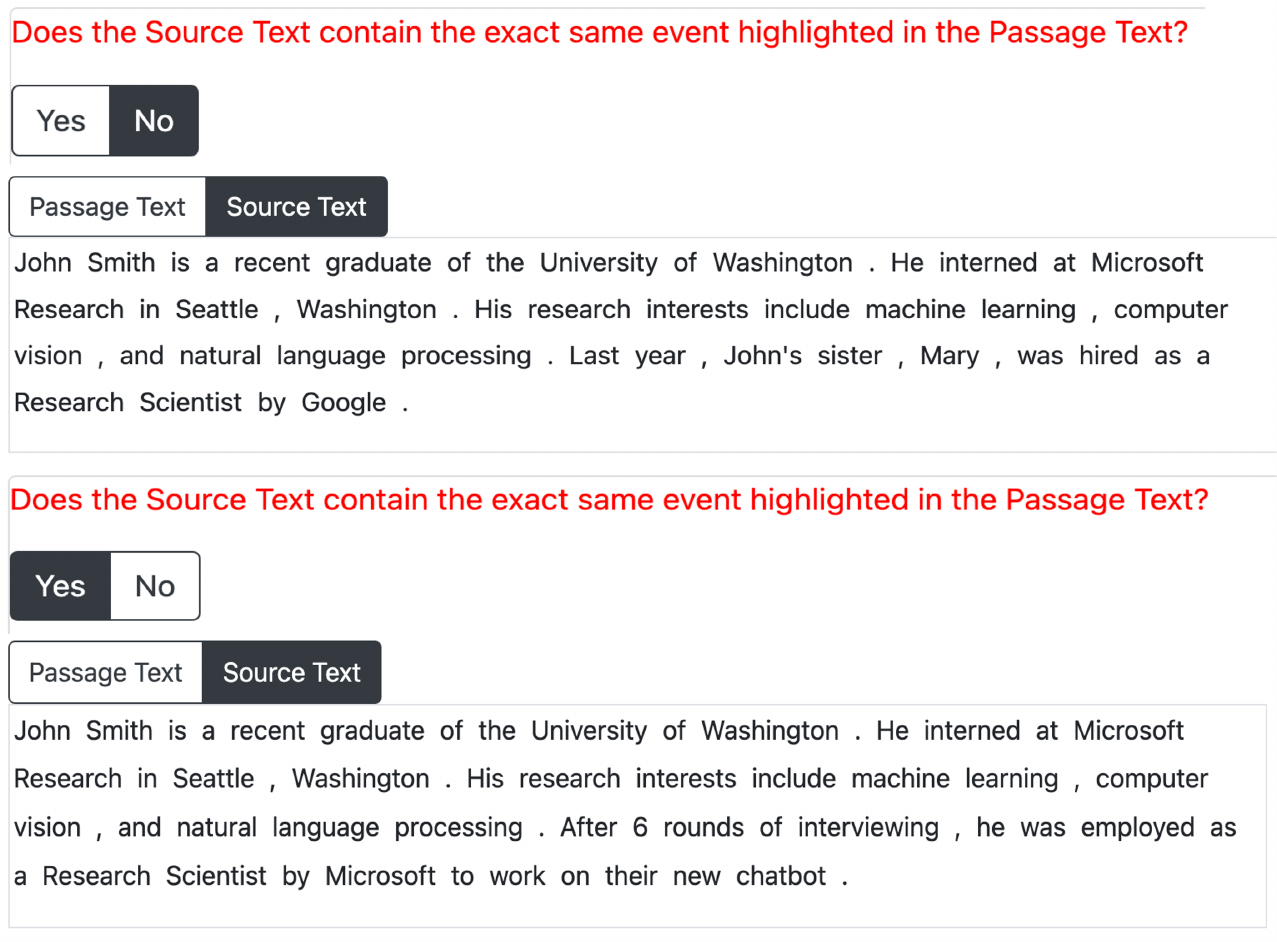}
    \caption{Two example source documents in the annotation protocol used for source validation. After reading the passage text (\autoref{fig:passage-val}) the annotators can switch to the source text and annotate either no (top) or yes (bottom) depending on the source document. In this figure the top document is annotated ``no" because the hiring event involves a hiring event at Google, but the bottom document is annotated ``yes" because the hiring event is present with the ``employed" trigger. These examples are constructed examples used in the instructions of the annotation protocol and not from MegaWika.}
    \label{fig:source-val}
\end{figure}

We evaluate the quality of source documents by validating if the situation described in the passage text is contained in the source document. To do this we provide qualified annotators with the annotation protocol in \autoref{fig:passage-val} and \autoref{fig:source-val}. In this protocol, the annotators are first shown the Wikipedia passage (\autoref{fig:passage-val}) that contains a highlighted situation trigger identified by the FrameNet parser from \cite{xia-2021-lome}. The annotator can then switch to the source passage (\autoref{fig:source-val}), where they will be shown the full source document. The annotator then must answer the question "Does the Source Text contain the exact same event highlighted in the Passage Text?". The annotators are instructed to annotate ``yes" even if the source text does not contain the same trigger used in the event. Each of the 3,504 passage source report pairs are annotated with 3-way redundancy and reports a Krippedorf's $\alpha$ of 0.4144 and taking the majority agreement of each document, we find that 48.23\% of the source documents contain the same event described in their corresponding Wikipedia passage.

 \newpage
 \section*{Appendix D}

The counts and percentages of the passages and their sentences that have been re-translated by Google Translate are presented in Table \ref{tab:retrans-stats}. For the languages in the left-hand side table, the re-translated passages have also been aligned with their original passages.

\begin{table}[h]
\begin{tabular}{|l|l|l|l|l|l|l|}
\cline{1-3} \cline{5-7}
lang & passage cnt. (pct.) & sentence cnt. (pct.) &  & Lang & passage cnt. (pct.) & sentence cnt. (pct.) \\ \cline{1-3} \cline{5-7} 
xh   & 816 (100)           & 1,436 (100)           &  &   hr   &          391,644 (100)           &     770,239 (100)                 \\ \cline{1-3} \cline{5-7} 
km   & 9,974 (100)          & 19,494 (100)          &  &    bn  &            466,766 (100)         &   705,390 (100)                   \\ \cline{1-3} \cline{5-7} 
ps   &     12,142 (100)     &      22,103 (100)     &  &   th   &      530,036 (100)               &     925,970 (100)                 \\ \cline{1-3} \cline{5-7} 
af   &   14,659 (100)     &   27,152 (100)    &  & ro     &      687,778 (100)         & 1,278,143 (100)                      \\ \cline{1-3} \cline{5-7} 
ga   &   20,603 (100)  &   33,872 (100)   &  &   ko   &    782,459 (95.1)                 &    1,497,452 (92.8) 
\\  \cline{1-3} \cline{5-7}
mn   &  23,528 (100)   &  47,757 (100)    &  &  cs    &     1,092,923 (100)                &     2,116,490 (100)
\\  \cline{1-3} \cline{5-7}
si   &   37,539 (100)  &  92,290 (100)    &  &   fa   &      1,112,258 (100)    &   1,898,707 (100)  
\\  \cline{1-3} \cline{5-7}
gu   &  44,279 (100)   &  68,528 (100)    &  &  fi    &        107,891 (9.4)             &   200,339 (9.6)  
\\  \cline{1-3} \cline{5-7}
ne   &  57,176 (100)   &  86,712 (100)    &  & nl     &       1,382,721 (100)              &     3,119,317 (100)
\\  \cline{1-3} \cline{5-7}
mr   &  58,434 (100)   &   115,736 (100)   &  &  id    &   1,556,901 (100)        &     2,842,698 (100)
\\  \cline{1-3} \cline{5-7}
my   & 96,517 (100)    &   146,139 (100)   &  &   tr   &        1,672,958 (100)             &     3,213,869 (100)
\\  \cline{1-3} \cline{5-7}
et   &  166,789 (100)   &  270,793 (100)    &  &   vi   &    2,306,144 (100)                 &   3,938,586 (100)  
\\  \cline{1-3} \cline{5-7}
lv   &  167,396 (100)   &   326,898 (100)   &  &   uk   &                2,279,129 (100)     &     4,406,834 (100)
\\  \cline{1-3} \cline{5-7}
kk   &  178,624 (100)   &   246,731 (100)   &  &  ja    &     132,110 (5.4)                &     291,090 (5.4)
\\  \cline{1-3} \cline{5-7}
sl   &  233,944 (100)   &   410,855 (100)   &  & zh     &     313,281 (12.0)                &   941,386 (14.1)  
\\  \cline{1-3} \cline{5-7}
lt   & 236,630 (100)    &  463,416 (100)    &  &  pl    &               1,968,266 (59.7)      &     3,523,356 (53.6)
\\  \cline{1-3} \cline{5-7}
ml   & 247,043 (100)    &  549,396 (100)   &  &  sv    &  95,302 (2.8)                   &     190,300 (3.4)
\\  \cline{1-3} \cline{5-7}
ka   &  247,213 (100)   & 411,145 (100)  &  &   it   &    3,207,857 (91.5)                 &     5,383,533 (87.6)
\\  \cline{1-3} \cline{5-7}
ur   & 257,172 (100)    &   370,496 (100)   &  &  pt    &                257,169 (5.9)     &     467,795 (5.6)
\\  \cline{1-3} \cline{5-7}
hi   &  264,269 (100)   &  395,372 (100)    &  &  ru    &  95,434
(2.0)          &    157,292 (1.8)
\\  \cline{1-3} \cline{5-7}
he   &   273,668 (100)  &  480,253 (100)    &  &   fr   &             2,648,489 (52.0)        &     5,509,544 (51.3)
\\  \cline{1-3} \cline{5-7}
mk   &  289,279 (100)   &   763,893 (100)   &  & ar     &                115,598 (2.2)     &    239,266 (1.8) 
\\  \cline{1-3} \cline{5-7}
az   &  328,181 (100)   &  597,600 (100)    &  &  en    &      n/a               &    n/a 
\\  \cline{1-3} \cline{5-7}
ta   & 344,012 (100)    & 620,087 (100)     &  &  de    &     177,736 (2.9)               &     297,550 (2.2)
\\  \cline{1-3} \cline{5-7}
gl   &   369,034 (100)  &   70,7823 (100)   &  & es     &  2,868,149 (46.3)        &     4,689,934 (46.6)
\\  \cline{1-3} \cline{5-7}
\end{tabular}
\caption{Statistics of the data with completed improved translations.}
\label{tab:retrans-stats}
\end{table}

\end{document}